\documentclass[letterpaper, 10 pt, conference]{ieeeconf}

\IEEEoverridecommandlockouts
\RequirePackage{scrlfile} 
\PreventPackageFromLoading{subfig}
\usepackage{cite}
\usepackage{amsmath,amssymb,amsfonts}
\usepackage{bm}
\usepackage{bbm}
\usepackage{graphicx}
\usepackage{textcomp}
\usepackage{xcolor}
\usepackage{subcaption}
\usepackage{verbatim}
\usepackage{tikz}
\usepackage{breqn}
\usepackage{booktabs}
\usepackage{multirow}
\usepackage{svg}
\usepackage{hyperref}
\usepackage{balance}
\usepackage{algorithm}
\usepackage{algpseudocode}

\newcommand{\realnum}{\mathbb{R}}

\def\BibTeX{{\rm B\kern-.05em{\sc i\kern-.025em b}\kern-.08em
    T\kern-.1667em\lower.7ex\hbox{E}\kern-.125emX}}
\begin{document}



\title{\LARGE \bf                                                                                        
Fully-probabilistic Terrain Modelling with Stochastic Variational Gaussian Process Maps                                      
}

\author{Ignacio Torroba$^{1}$  Christopher Iliffe Sprague$^{1}$ and John Folkesson$^{1}$
\thanks{$^{1}$The authors are with the \href{https://smarc.se/}{Swedish Maritime Robotics Centre (SMaRC)} and the \href{https://www.kth.se/rpl/division-of-robotics-perception-and-learning}{Division of Robotics, Perception and Learning} at KTH Royal Institute of Technology, SE-100 44 Stockholm, Sweden                                          
        {\tt\small \{torroba, sprague, johnf\}@kth.se}}%
}

\maketitle                                                                                               
\thispagestyle{empty}                                                                                    
\pagestyle{empty}

\begin{abstract}
Gaussian processes (GPs) are becoming a standard tool to build terrain representations thanks to their capacity to model map uncertainty. This effectively yields a reliability measure of the areas of the map, which can be directly utilized by Bayes filtering algorithms in robot localization problems. A key insight is that this uncertainty can incorporate the noise intrinsic to the terrain surveying process through the GPs ability to train on uncertain inputs (UIs). However, existing techniques to build GP maps with UIs in a tractable manner are restricted in the form and degree of the input noise. In this letter, we propose a flexible and efficient framework to build large-scale GP maps with UIs based on Stochastic Variational GPs and Monte Carlo sampling of the UIs distributions. We validate our mapping approach on a large bathymetric survey collected with an AUV and analyze its performance against the use of deterministic inputs (DI). Finally, we show how using UI SVGP maps yields more accurate particle filter localization results than DI SVGP on a real AUV mission over an entirely predicted area. 
\end{abstract}


\section{Introduction}
Seabed surveying with sonar-equipped, autonomous underwater vehicles (AUVs), as opposed to surface vessels, provides higher-resolution bathymetric maps due to a closer proximity of the AUV to the sea bottom. This higher resolution, however, comes at the expense of a cumulative drift in the vehicle's dead reckoning (DR) estimate due to the lack of GPS signal underwater. 
Among sonars, multibeam echosounders (MBES) have become standard in bathymetric surveying thanks to the ease of interpreting the 3D data they provide, which can be treated through existing point cloud manipulation techniques \cite{torroba2020pointnetkl}. However, they have limited resolution and the beam-forming process is inherently prone to noise. Due to this, bathymetric point cloud maps tend to be memory-intensive, noisy and have gaps in the coverage. 

Gaussian processes (GPs) have been demonstrated to be an invaluable technique to tackle some of these problems when building maps, as they can regress the original seabed structure from the noisy, sparse MBES beams collected during the survey. The result is a lighter representation of the terrain when compared to raw point clouds, which naturally smooth out MBES noise and can predict unseen areas conditioned on the collected data. Furthermore, being a Bayesian inference technique, GPs provide a measure of the variance of the resulting map. This results in maps which provide a reliability measure of the terrain modelled, from which localization algorithms based on Bayes filters can benefit directly.

Another capability of GPs is that they can fuse the uncertainties emanating from the vehicle's DR and the sensor used in the data collection into their learning process. These can be propagated into the sonar beams' uncertainty to create uncertain inputs (UIs), which the GP can learn. The work in \cite{girard2004approximate} shows how the use of UIs yields more robust predictions.
The most common technique currently used for training GP maps with UIs was introduced in \cite{o2012gaussian} and has since been applied in several works in the field of autonomous mapping and informative path planning \cite{popovic2020informative, ghaffari2019sampling}.
This method is based on fusing the UIs uncertainties directly into the GP kernel through a Gauss-Hermite quadrature of the covariance function. However it is limited by the need to manually adjust the number of sampling points $M$ for the quadrature to adequately approximate the degree of uncertainty in the data. Moreover, this formulation imposes the restriction on the UIs to be Gaussian.

In this work, we propose as an alternative the use of Monte Carlo (MC) sampling of the UIs together with a Stochastic Variational Gaussian Process (SVGP) \cite{hensman2013gaussian} in order to propagate the input uncertainties into the map. We show that this approach can handle large amounts of training data, and neither imposes any constraint on the UIs distribution nor requires any prior knowledge on the degree of uncertainty in the inputs. We test this method on a very large bathymetric survey with an AUV and measure the advantages of the use of UIs against deterministic inputs (DIs) for learning SVGP maps in the tasks of bathymetry reconstruction and prediction. Finally, to show the amenability of SVGPs with UIs to practical applications, we demonstrate AUV localization on an entirely predicted area. We compare the results against localization on DI SVGP maps and show how the UIs SVGP predictions result in a more accurate AUV pose estimate. The framework developed for this work has been released and can be found in this repository\footnote{\url{https://github.com/ignaciotb/UWExploration/}}.

\section{Related work}
Gaussian processes are a non-parametric, Bayesian regression method capable of modelling real-world problems in a fully probabilistic manner.
Originally, their main limitations was their cubic scalability with the number of training inputs and their limited applicability only to Gaussian likelihoods \cite{seeger2004gaussian}. Regarding scalability, the introduction of a sparse formulation based on a low-rank approximation to the kernel \cite{williams2001using} granted a reduction both in computational complexity and memory requirements of the GP implementations \cite{csato2002sparse}. Regarding applicability, the adaption of variational inference techniques to learning GP models in \cite{titsias2009variational} allowed posing the training as an optimization problem and  approximating non-Gaussian GP likelihoods. Building upon this approach, \cite{hensman2013gaussian} presented a stochastic minibatch optimization framework for GP training capable of handling datasets of hundreds of thousands of samples, the Stochastic Variational GP.

Concerning the treatment of noisy training data, the GP formulation originally assumed the input locations to be deterministic (DIs) in order to make the training tractable. This, however, hindered its application to many real-world problems that dealt with noise-corrupted inputs. In \cite{girard2004approximate}, Girard et al. derived a closed-form formulation for GP kernels that accounted for Gaussian uncertainty in the inputs. Although successful in fusing the inputs uncertainty into the GP training, this approach was only applicable to a few amenable Gaussian kernels. The later heteroscedastic approaches \cite{le2005heteroscedastic, mchutchon2011gaussian} would link the inputs' variances to the outputs proportionally through the local, approximated linear gradient of the GP posterior mean.

In the field of autonomous terrain modelling, GPs with UIs have been applied to map regression from noisy sensory inputs in several contexts. Following the conclusions from \cite{girard2004approximate}, in \cite{o2012gaussian} O'Callaghan et al. employed a Gaussian-Hermite quadrature approximation to the GP kernel to fuse the UIs' uncertainty into the GP training in a tractable manner. They showed the advantages of using UIs in a grid mapping task with a mobile robot equipped with a laser scanner. However, this approximation required the UIs to be Gaussian and manually specifying the number of samples for the quadrature a priori, which depends on the degree of uncertainty in the UIs. This would necessitate prior knowledge of the training data and a similar degree of uncertainty across the whole dataset for an optimal number of samples to be found. Besides, quadrature approximations are known to scale poorly with the dimensions of the input data. Despite these shortcomings, this technique is still being used in the latest works in the field \cite{ghaffari2019sampling, popovic2020informative}.

SVGPs have been applied to map modelling both in the land \cite{senanayake2017learning} and the underwater domains \cite{chen2019long}. Their preferred use over other GP modalities was justified by the need for a fast learning procedure capable of handling large datasets in an online fashion. However, in both cases the inputs were assumed to be noise-free and therefore DIs were used.
To the best of our knowledge, this work is the first to present results on SVGP mapping with UIs by leveraging the minibatch optimization technique for SVGP training with the unbiased MC approximation of the distribution of the uncertain inputs.

Finally, there are several instances of vehicle localization on GP maps based on particle filters (PF) in the literature. Most notably, in \cite{ko2009gp} both the prediction and measurements functions are modelled through GPs. In the underwater domain, \cite{zhou2016terrain} presents an AUV PF localization framework in a GP map and compares its performance in simulation against different bathymetry reconstructions. Going beyond the aforementioned works, in this letter we present the results of a real AUV PF localization test mission on an entirely predicted area of seabed on the SVGP map.

\section{SVGP bathymetric maps with UIs}
In this section, our approach to learn SVGP maps with UIs is presented. Although we describe the method with the view on bathymetric mapping with AUVs and sonar sensors, the framework is general and could be applied to any equivalent instance of autonomous terrain construction under localization and sensor uncertainty.

\label{sec:svgp_mapping}
\subsection{Autonomous Bathymetric Surveying}
Given an AUV equipped with a 3D mapping sensor, the aim is to autonomously reconstruct an unknown area of seabed.
The first-order Markov dynamics of the 6D AUV state at time $t$ is given by $r_t = g(r_{t-1}, c_{t})$ where $c_t$ is the control input to the motion model. The MBES measurement model $p_{i} = h(r_{t}, e_{i})$ provides a 3D sample of terrain $p_i = [x, y, z]$ from the patch $e_{i}$ where the beam $i$ hits the seabed for a given vehicle pose $r_t$. If we model the inherent noise in the real DR and measurement models as white Gaussian with $\nu_{t} \sim \mathcal{N}(0, W)$ and $\delta_{t} \sim \mathcal{N}(0, Q)$ respectively (with $W \in \realnum^{6 \times 6}$ and $Q \in \realnum^{3 \times 3}$) both $r_{t}$ and $p_{i}$ follow probability distributions given by:
\begin{align}
\label{eq:dr_and_h}
    r_{t} &\sim \mathcal{N}(g(r_{t-1}, c_{t}), W) \\
    P_{t} &\sim \mathcal{N}(h(r_t, E_t), Q),
\end{align}
where the measurement model has been generalized to gather the $n$ beams collected in the same ping $t$, $P_t = \{p_{i}\}_{i=1}^{n}$ corresponding to a set of patches of seabed $E_t = \{e_{i}\}_{i=1}^{n}$ in the map frame. 

After a survey in which $N$ sensor observations or samples from the terrain have been collected into a 3D point cloud, a dataset of the form $D = \{p^{xy}_i, p^{z}_i\}_{i=1}^{N}$ can be constructed. In the remainder of the paper we will also use the notation $D = \{X, Y\}$ in place of the previous definition to follow the standard notation in GP literature.

\subsection{Stochastic Variational Gaussian Processes Maps}
The dataset $D$ can be used to regress a function of the form $f: X \to Y$ with a SVGP, which will produce a continuous representation of the underlying terrain from which the samples have been collected. Additionally, the posterior of $f(X)$ on the observations will allow us to perform predictions on unseen data conditioned on $D$.
Applying Bayes theorem this posterior can be computed as:
\begin{equation}
\label{eq:bayesian_svgp}
    p(f(X) | Y) = \frac{p(Y | f(X)) p(f(X))}{p(Y)}
\end{equation}
We choose our prior to be a Gaussian process such that $p(f(X)) \sim  \mathcal{N}(0, \mathcal{K}_{NN})$ modelled by its kernel $\mathcal{K}_{NN}$, with a set of hyperparameters $\theta$. However, computing that posterior is intractable due to the marginal likelihood $p(Y)$ and therefore SVGP applies variational inference to approximate $p(f(X) | Y) \approx q(f(X))$. Where $q(f(X)) \sim  \mathcal{N}(\tau, \beta)$ is a Gaussian variational term.

Given a specific model, the optimal $\theta$ can be learned maximizing the likelihood of observing our $N$ collected terrain samples for that given model. In variational inference this is equivalent to maximizing the ELBO of Eq. \ref{eq:bayesian_svgp} with respect to the model parameters. The ELBO is derived minimizing the KL divergence between our target posterior and the variational term: $\text{argmin}_{\theta} KL [q(f(X)|| p(f(X)|Y)]$. However, this optimization involves the inversion of the very large kernel matrix $\mathcal{K}_{NN}$. In order to overcome this, sparse GPs formulation makes use of a low-rank approximation to the kernel $\mathcal{K}_{NN} \approx \mathcal{K}_{NS}\mathcal{K}_{SS}^{-1}\mathcal{K}_{SN}$ using $S << N$ inducing variables $u_s = f(Z_s)$ in the model. $Z_s$ are the locations of the inducing points, which belong in the same domain as $X$ in this work. With the use of the inducing variables, the new target KL divergence to optimize becomes $KL [q(f(Z)|| p(f(Z)|u)]$, which can be approximated using Jensen's inequality:
\begin{multline}
\label{eq:elbo_svgp}
    \mathcal{L} = \sum^N_i\mathbb{E}_{q(f(x_i))} \left[\log p(y_i \mid f(x_i))\right] - \\ \mathrm{KL}\left[q(u) \mid\mid p(u)\right]
\end{multline}
All the components in the ELBO above are Gaussian and can be derived.
Therefore, standard stochastic gradient descent (SGD) techniques can be applied to its optimization. This, together with the use of the inducing points formulation, allows us to handle very large training datasets through the use of minibatch optimization.
This is another key reason to advocate for the use of SVGPs when constructing large bathymetric maps containing hundreds of thousands of samples.

Finally, the ELBO in Eq. \ref{eq:elbo_svgp} can be derived with respect to the locations of the inducing points $Z$ and so these can be added as hyperparameters within $\theta$ and learned jointly.
Despite adding computational workload to the training, learning the optimal location of the inducing points will result in a more accurate representation of the bathymetry \cite{titsias2009variational}. Thus, it should always be considered if the training time is not a constraint (such as in an offline context).  

\subsection{Vehicle Uncertainty Propagation to the SVGP Map}
So far we have assumed that our training data is deterministic. But for a fully probabilistic approach to autonomous mapping, all the system uncertainties in the process need to be taken into account. There are two main sources of noise in surveying: those emanating from the vehicle's DR and those related to physical sensors.
Thus, we first introduce how the AUV pose estimate and MBES uncertainties, modelled in Eq. \ref{eq:dr_and_h}, are propagated to the beams and fused. After that, we show how standard SVGP can be readily used to handle the resulting UIs through the use of MC integration.

If we now redefine the training inputs as samples from the terrain corrupted with Gaussian noise, we obtain:
\begin{equation}
    p_i = \bar{p}_i + \epsilon_{pi} \;\; \text{with} \;\;  \epsilon_{pi} \sim \mathcal{N}(0, \Sigma_i)
\end{equation}
Through this $p_i \sim \mathcal{N}(\bar{p_i}, \Sigma_i)$ is a Gaussian distribution and we can model through $\Sigma_i$ the uncertainty propagated from the data collection process.
We compute $\bar{p_i}$ and $\Sigma_i$ through the sigmapoint method originally presented in \cite{julier1996general}. We choose to do so following the results in \cite{barfoot2014associating}, which argues that this method provides the most accurate results for uncertainty propagation through a nonlinear measurement model. We implement our uncertainty propagation framework with the tools to manipulate poses and associated uncertainties presented in \cite{barfoot2014associating}.
The reader is referred to \cite{barfoot2014associating} for a full explanation of the method.

For every DR pose estimate in the AUV trajectory $\bar{r_t}$, and the $n$ seabed patches observed from that pose, $e_{i,t} \sim \mathcal{N}(\bar{e}_{i,t},\Omega)$, a covariance matrix $\Xi_t \in \realnum^{9 \times 9}$ is compounded from $W_t$ (Eq. \ref{eq:dr_and_h}) and $\Omega$. Then, the $l$ sigma point samples $\{r_l, e_l\}$, which are used to approximate the mean and variance of each measurement $\{\bar{p}_{i}, \Sigma_{i}\}$, are computed as follows:
\begin{align}
\label{eq:sigmapoint}
    CC^T &= cholesky(\Xi_t) \\
    \lambda_l &= 0 \;\; l = 0\\
    \lambda_l &= \sqrt{L + \kappa} \: col_l C, \;\; l = 1,...,L \\
    \lambda_{l+L} &= -\sqrt{L + \kappa} \: col_l C, \;\; l = 1,...,L\\
    \begin{bmatrix} 
            \xi_{l} \\ \zeta_{l} 
    \end{bmatrix} &= \lambda_l \\
    r_l &= exp(\xi^{\wedge}_{_l})\bar{r_t}\\
    e_l &= \bar{e_i} + D\zeta_l,
\end{align}
where $col_l$ is the $l$ column of the matrix, $\xi_l \in \realnum^6$ and $\zeta_{l} \in \realnum^3$ are the perturbations applied to the AUV pose and seabed patches respectively, $\wedge$ and $exp()$ are the operators to turn them back and forth into $\realnum^{4 \times 4}$ matrices belonging to the $SE(3)$ group, $D$ is a dilation matrix and $L=9$ and $\kappa = 0$.
Each sample is passed through the nonlinear MBES model in Eq \ref{eq:dr_and_h}, $p_l = h(r_l, e_l)$ with $l = 0,..., 2L$ and the desired covariances for each UI are computed as follows:
\begin{align}
\label{eq:sigma_results}
    \bar{p}_i &= \frac{1}{L+\kappa} \left(\kappa p_0 + \frac{1}{2} \sum_{l=1}^{2L} p_l\right) \\
    \begin{split}
        \Sigma_{i} &=  \frac{1}{L+\kappa}  \left(\kappa (p_0 - \bar{p}_i) (p_0 - \bar{p}_i)^T \right. \\
        &+ \left. \frac{1}{2}\sum_{l=1}^{2L}(p_l - \bar{p}_i)(p_l - \bar{p}_i)^T\right)
    \end{split}
\end{align}

Considering now the training data as distributions instead of samples, our dataset becomes $D = \{\langle p^{xy}_i, \Sigma^{xy}_i \rangle, p^{z}_i\}_{i=1}^{N}$, or equivalently $D = \{X, Y\}$ with $X\sim \mathcal{N}(p^{xy}_i, \Sigma^{xy}_i)$.
The way the SVGP can handle these UIs is once again thanks to its capacity to operate on large datasets, together with the approximation capabilities of Monte Carlo sampling: Now for every learning step, a minibatch is constructed with $m << N$ deterministic training samples drawn from the $m$ randomly selected UIs distributions $X_i$ in a Monte Carlo fashion (one sample per UI, per minibatch iteration). As the SVGP training progresses, the MC sampling approaches the UIs original distributions while propagating their noise into the learning process through stochastic sampling. Since the incoming data in SVGP is ultimately of deterministic form, the SVGP does not need any reformulation to handle UIs and therefore any standard implementation can be used for this approach. 

\section{Particle Filter Localization on SVGP Maps}
This section briefly presents the application of SVGP maps to the task of vehicle localization with a particle filter. 
PF localization on GP maps requires an accurate terrain reconstruction but also well-callibrated map variances for a correct vehicle pose estimate, since the GP-PF utilizes both. We argue that our method's harnessing of the uncertainty in the map creation process provides better estimates of both parameters, resulting in a more accurate PF pose estimate than through the use of DI SVGP maps.

\subsection{The SVGP-PF Algorithm}
A particle filter approximates the posterior distribution of a vehicle pose estimate through a set of $J$ finite sample states given by Eq. \ref{eq:particles}, following Algorithm \ref{alg:gp_pf}.
\begin{equation}
    \label{eq:particles}
    R_t = \{\langle r^j_t, w^j_t \rangle | j = 1, ..., J\}
\end{equation}

\begin{algorithm}
\caption{The SVGP Particle filter}\label{alg:gp_pf}
\begin{algorithmic}[1]
\Require$(R_t, c_{t, t-1}, P_t, Q_t):$  
\State $\hat{R_t} = R_t = \emptyset$
\For{$j = 1$ to $J$}
    \State $r_t^j \sim \mathcal{N}(g(r_{t-1}, c_{t}), W)$
    \State $\hat{P}_t^j = meas(r_{t}^j, P_t)$
    \State $w_t^j = \mathcal{N}(P^{z}_t; GP_{\mu}(\hat{P}_t^{j, xy}), GP_{\Sigma}(\hat{P}_t^{j, xy}) + Q_{P^z_t})$
    \State $add$ $\langle r^j_t, w^j_t \rangle$ $to$ $\hat{R}_t$
\EndFor
\For{$j = 1$ to $J$}
    \State $draw$ $i$ $with$ $probability$ $\propto w_k^i$
    \State $add$ $r_t^i$ $to$ $R_t$
\EndFor
\State \Return $R_t$
\end{algorithmic}
\end{algorithm}

In algorithm \ref{alg:gp_pf}, the function $meas$ in line 4 is a transformation of the real MBES ping at time $t$ to the pose of particle $j$ given by $r_t^j$.
The weight in line 5 models the probability of the difference between the depth components in the ping $t$ and those predicted by the SVGP for particle $j$ being zero. 
These predicted measurements can be computed through the SVGP predictive posterior from Eq. \ref{eq:bayesian_svgp} on query points $X_{*}$, which has the form:
\begin{multline}
\label{eq:posterior_svgp}
    p(f(X_{*}) | Y) \sim  \mathcal{N}(K_{*s}K^{-1}_{ss}\mu, \\
    (K_{*s}K^{-1}_{ss})\Sigma(K_{*s}K^{-1}_{ss})^T + K_{**} - K_{*s}K^{-1}_{ss}K^{T}_{*s}))
\end{multline}
Having $X_{*} = \hat{P}^{j,xy}$, line 5 can be formulated as $w_t^j = p((P^z_t - GP(\hat{P}_t^{j, xy})) = 0)$. This probability distribution is normal and naturally utilizes both the $z$ component of the noise model of the pings $Q_{P^z_t}$ and the uncertainty predicted by the SVGP $GP_{\Sigma}(\hat{P}_t^{j, xy})$.

\section{Experiments}
The experiments below compare the performance of SVGP maps trained with UIs with the presented method against DIs in the tasks of map modelling and prediction for different levels of noise in the system. Additionally, PF localization experiments are carried out over predicted terrain to assess the performance of UI SVGP against DI SVGP maps on a real application.

\subsection{The Bathymetric Dataset}
\begin{figure}[htbp]
    \centering
    \includegraphics[width=\linewidth]{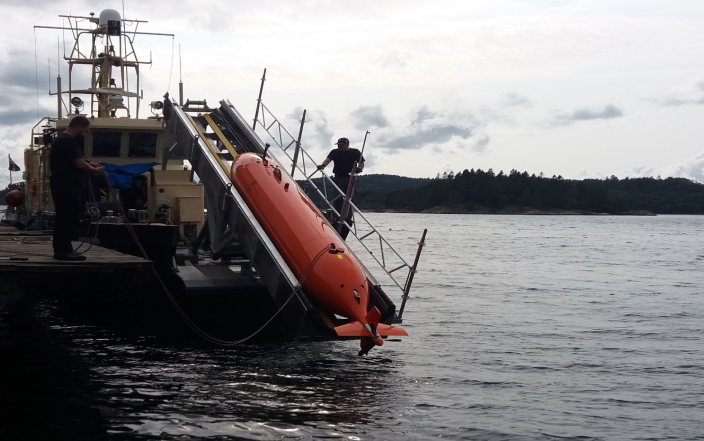}
    \caption{Hugin AUV being deployed on the survey area}
    \label{fig:hugin}
\end{figure}

\begin{figure*}[!ht]
\vspace*{0.1in}
\hspace{.6in}(a)\hspace{1.4in}(b)\hspace{1.4in}(c)\hspace{1.4in}(d)\\
\centering
\begin{subfigure}[]{.035\linewidth}
\centering
		\includegraphics[width=\linewidth]{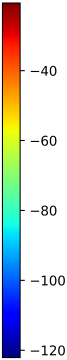}
\end{subfigure}
\begin{subfigure}[]{.22\linewidth}
\centering
	\includegraphics[width=\linewidth, trim=4 4 4 4,clip]{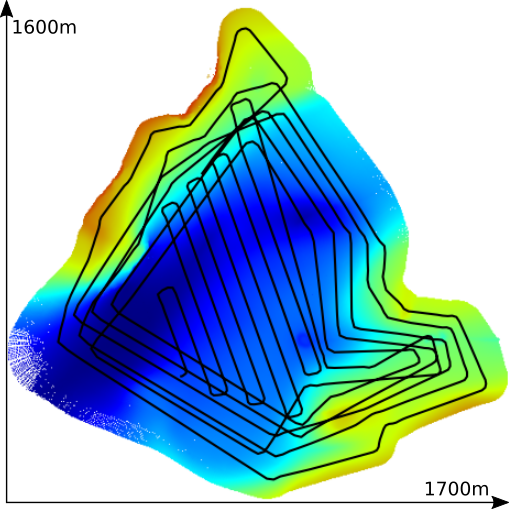}
\label{fig:pcl_map}
\end{subfigure}
\begin{subfigure}[]{.21\linewidth}
\centering
        \includegraphics[width=\linewidth]{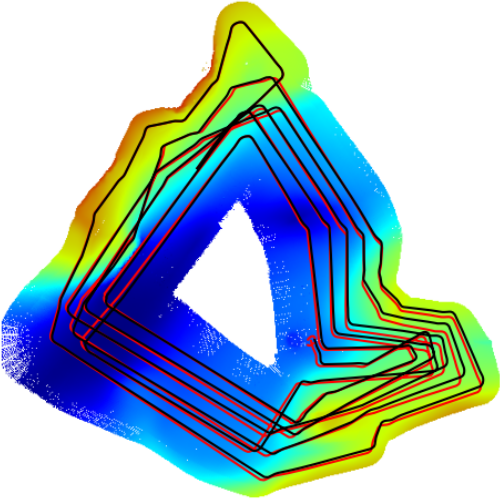}
\label{subfig:disrupted_bathy}
\end{subfigure}
\begin{subfigure}[]{.21\linewidth}
\centering
		\includegraphics[width=\linewidth]{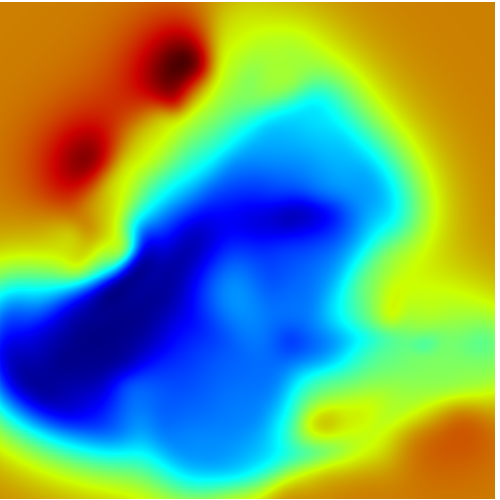}
\label{subfig:gp_ui}
\end{subfigure}
\begin{subfigure}[]{.21\linewidth}
\centering
		\includegraphics[width=\linewidth]{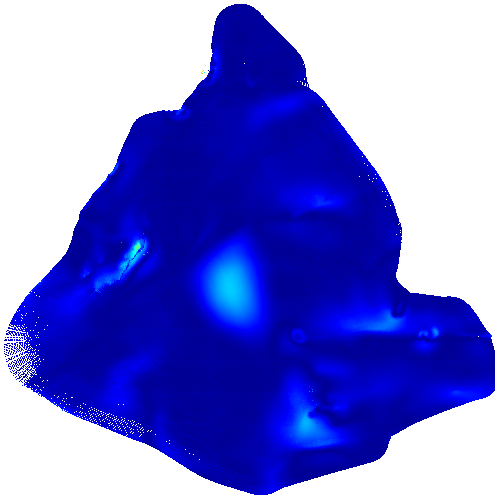}
\label{subfig:rmse_gp_ui}
\end{subfigure}
\begin{subfigure}[]{.03\linewidth}
\centering
		\includegraphics[width=\linewidth]{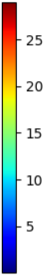}
\end{subfigure}
\caption{Left to right: a) full GT bathymetry, b) corrupted DR point cloud, c) UI SVGP posterior mean trained with b), and d) consistency error map between b) and the posterior mean in c) sampled.}
\label{fig:gp_maps_resuls}
\end{figure*}

For the experiments, a bathymetric survey was collected off the Swedish west coast with a MBES Kongsberg 2040 mounted on an AUV Hugin 3000, following the procedure explained in Section \ref{sec:svgp_mapping}.
The raw bathymetric point cloud can be seen in Fig. \ref{fig:gp_maps_resuls} a) and Fig. \ref{fig:hugin} shows the Hugin being deployed on the area.
The full mission lasted approximately 12.7 hours and covered a trajectory of 30.5 $km$ at an average speed of 1.5 $m/s$ .
The resulting dataset $D$ contains approximately 4.038.000 MBES beams over an area of approximately 1.4 $km^2$.

\subsection{The SVGP Training}
The SVGP maps have been implemented within the Gpytorch framework \cite{gardner2018gpytorch}.
A Matern covariance function \cite{rasmussen2003gaussian} with a smoothness parameter of $\tau = 1/2$ was selected as the SVGP kernel, as in \cite{sprague2020learning}.
The optimizer applied has been Adam \cite{kingma2014adam} with a learning rate of $1e^{-1}$ and a minibatch size of $m=4000$ points per iteration.
The same exponential moving average over the ELBO \cite{balandat2019botorch} has been used as a terminating criterion for the optimization of SVGPs with DIs and UIs.
The number of inducing points has been manually set to $S=1000$ based on prior knowledge of the dataset and the capabilities of the hardware. Existing research on methods to include this number as a hyperparameter in the learning process \cite{galy2021adaptive, senanayake2017learning} should be considered in an online setup.
Finally, the SVGPs training has been carried out in a laptop with an NVIDIA GeForce GTX 1060 GPU.

\subsection{SVGP Mapping Evaluation}
In order to directly compare the SVGP maps trained with DIs and UIs against the real bathymetry, we would need access to the ground truth map of the seabed. Since this is generally not an option in underwater surveying, for our experiments we have used the collected data as ground truth. Although not ideal, this is a reasonable assumption based on the high quality of the DR system on Hugin and the MBES model used. In order to generate lower quality surveys, we have corrupted the vehicle DR with additive white noise and propagated the vehicle and sensor uncertainties to the resulting bathymetric point clouds for the GP training.

To assess the quality of the SVGPs trained under DR distortion when compared to the GT bathymetry, we use the method presented in \cite{roman2006consistency} and previously used in \cite{bore2018sparse}.
This method computes the consistency error between the original bathymetry and the point cloud generated from sampling the SVGP posterior on a grid over the survey area with $100^2$ samples.
We use the error between the GT map and the raw point cloud generated from the disrupted survey as the baseline error to compare the SVGP maps. This is because this error is intrinsic to the data collection process and cannot be eliminated through different map modelling techniques.

In order to evaluate both the reconstruction and prediction errors of the SVGPs with DI and UIs, we have divided the survey into two sections. The training dataset contains the outer loops of the survey, as seen in Fig. \ref{fig:gp_maps_resuls} b) and consists of 2.650.000 MBES beams. The SVGPs have been trained on this data and the reconstruction error has been measured directly comparing the SVGPs posteriors to that section of the raw GT point cloud. The second survey section contains the remaining MBES beams over the central area of the map. This subset has not been used for training, only for validation of the SVGP predictions over a large, unseen area. Thus, the prediction consistency error has been computed by comparing the SVGP posterior point clouds to the full GT bathymetry and subtracting the reconstruction error.

\begin{table*}[!t]
\vspace*{0.1in}
\caption{Average results comparing the performance of SVGP mapping with DIs and UIs.} \label{tab:maps_results}
\centering
\begin{tabular}{|c|ccc|cc|cc|cc|}
\hline
\multirow{2}{*}{\shortstack[l]{DR iid yaw noise (rad/s)}} & \multicolumn{3}{c|}{Reconstruction RMSE (m)} & \multicolumn{2}{c|}{Prediction RMSE (m)} & \multicolumn{2}{c|}{Learning steps} & \multicolumn{2}{c|}{Tr($K_{ss}$)} \\ \cline{2-10} 
                  & \multicolumn{1}{c|}{DR PCL} & \multicolumn{1}{c|}{SVGP Di} & \multicolumn{1}{c|}{SVGP Ui} & \multicolumn{1}{c|}{SVGP Di} & \multicolumn{1}{c|}{SVGP Ui} & \multicolumn{1}{c|}{SVGP Di} &        \multicolumn{1}{c|}{SVGP Ui} & \multicolumn{1}{c|}{SVGP Di} & \multicolumn{1}{c|}{SVGP Ui} \\ \hline
                  0 & 0.750 & 1.226 & 1.179 & 1.660 & 1.678 & 8014.0 & 8015.0 & 58375.59 & 59230.035 \\ 
                  $1e^{-3}$ & 2.354 & 2.461 & 2.437 & 0.766 & 0.566 & 8512.6 & 9045.8 & 60001.240 & 61666.679 \\ 
                  $2e^{-3}$ & 3.366 & 3.309 & 3.308 & 0.697 & 0.554 & 8859.4 & 9189.2 & 66936.702 & 68230.707 \\ \hline
\end{tabular}
\end{table*}

\subsection{Experimental Setup}
\begin{figure}[htbp]
    \centering
    \includegraphics[scale=0.22]{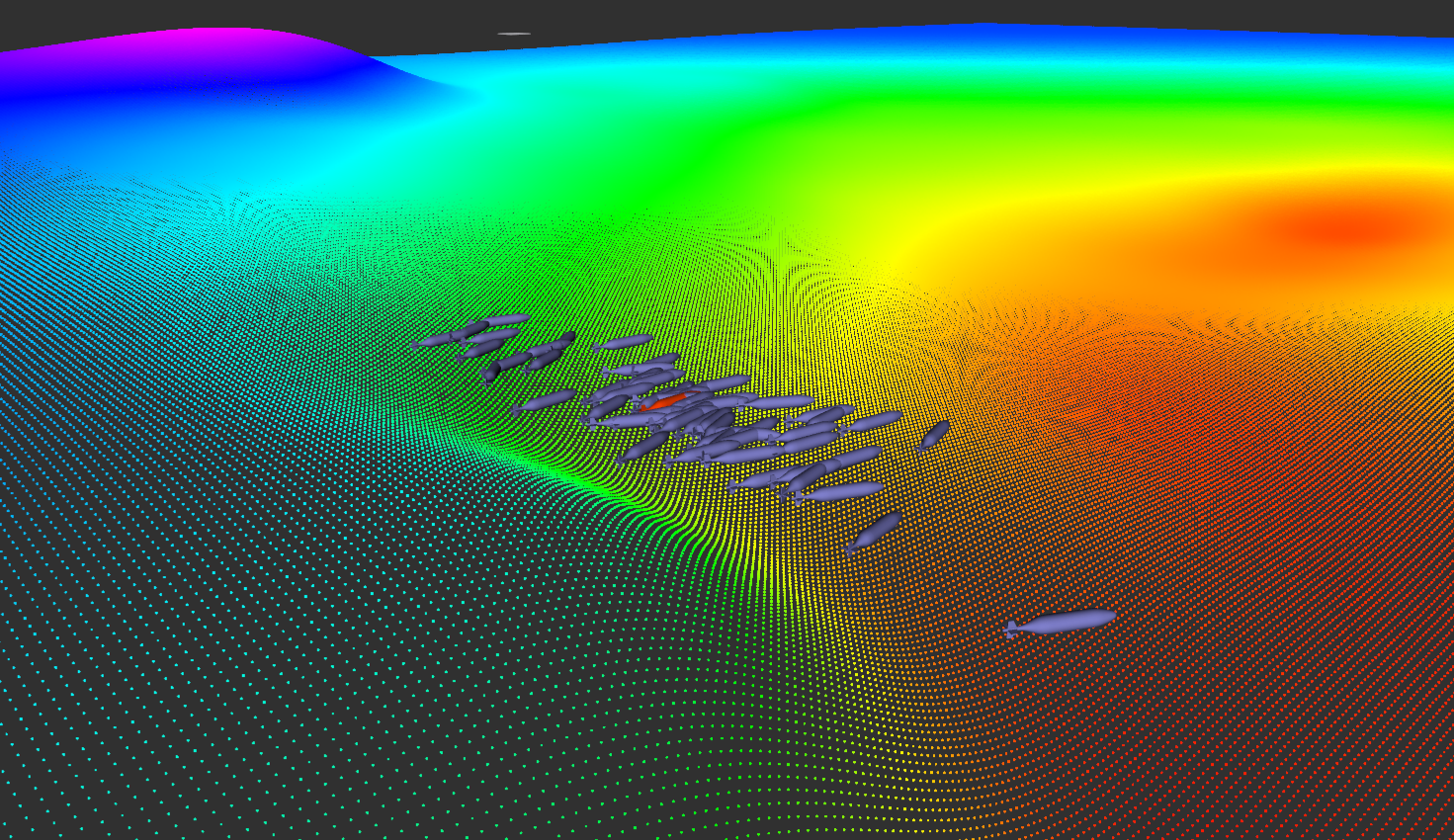}
    \caption{Visualization of the AUV PF filter in a SVGP map}
    \label{fig:auv_env}
\end{figure}

The framework used to replay, disrupt and visualize the AUV mission for the uncertainty propagation and PF experiments has been implemented in ROS. Figure Fig. \ref{fig:auv_env} shows the visualization in RViz of the PF running on a SVGP map. The GT Hugin pose is marked by the orange marker, with the particles represented in grey.

Since we do not have access to the original Hugin localization uncertainty, this has been approximated when generating the training datasets. The DR uncertainty has been computed through an EKF with a 6 DOF point-mass motion model, using as inputs the interpolated AUV velocities between time steps.
Regarding the PF experiments, the actual implementation of Eq. \ref{eq:posterior_svgp} of the SVGP measurement model omits the computational loop in Algorithm \ref{alg:gp_pf} to improve the computational complexity. Namely, instead of executing $J$ individual calls to the GPU to compute the particles expected measurements, all the query points are packed and sent together, optimizing GPU memory allocation.
Finally, the method used in line 9 of Algorithm \ref{alg:gp_pf} is the residual resampling, introduced in \cite{liu1998sequential}.

\section{Results}
This section presents and analyzes the results from the experiments described above on the tasks of terrain modelling and AUV PF localization.

\subsection{Comparison of SVGP Mapping Performance}
Three instances of the training set have been generated differing on the noise level applied to corrupt the DR. Only the heading of the vehicle has been degraded, with the values used seen on Table \ref{tab:maps_results}.
The noise model of the prediction step of the EKF ($W$ in Eq \ref{eq:dr_and_h}) has been adapted proportionally to the yaw noise for the three cases. 
Due to the stochastic nature of both the dataset generation and the SVGP training, five corrupted datasets have been created per DR noise level in order to collect statistically representative results.

\begin{figure}[htbp]
\centering
\begin{subfigure}[]{.43\linewidth}
\centering
		\includegraphics[width=\linewidth]{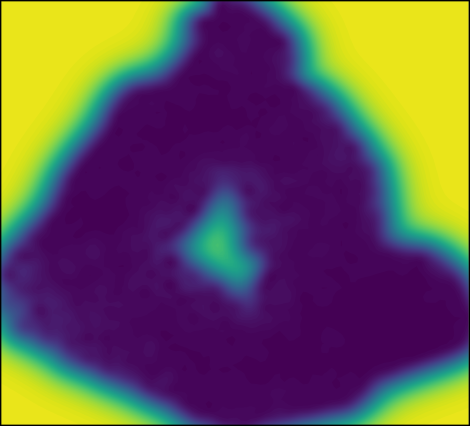}
\label{subfig:var_di}
\end{subfigure}
\begin{subfigure}[]{.43\linewidth}
\centering
		\includegraphics[width=\linewidth]{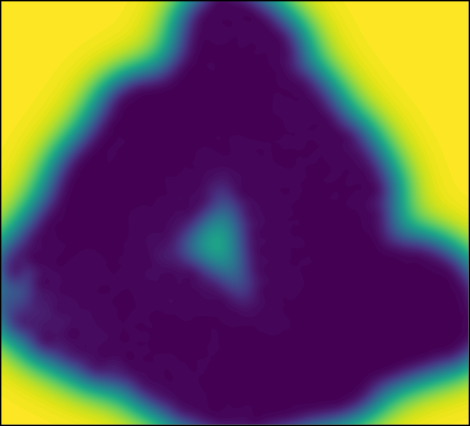}
\label{subfig:var_ui}
\end{subfigure}
\begin{subfigure}[]{.1\linewidth}
\centering
		\includegraphics[width=\linewidth, height=3.3cm, keepaspectratio]{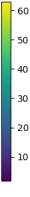}
\end{subfigure}
\caption{DI SVGPs (left) and UI SVGPs (right) posterior variances sampled over a grid.}
\label{fig:vars_gps}
\end{figure}

Fig \ref{fig:gp_maps_resuls} (b-d) show a disrupted bathymetric point cloud from the DR collection, the corresponding SVGP trained with UIs and the consistency error map between its posterior and the full GT bathymetry.
The average map consistency errors across the tests are summarized in Table \ref{tab:maps_results}. The RMSE values for different DR disruption levels indicate very similar reconstruction errors for both DI and UI SVGPs, with these errors being very close to the original DR point cloud errors. Interestingly, this suggests that the use of UIs does not influence greatly the mapping capacity of SVGPs. However, the prediction error, i.e. error in the area where we removed the original training data, shows a significant difference in performance between training methods. Indeed the SVGP UIs predicted terrain is consistently more accurate for both levels of DR disruption and almost equivalent for no disruption. 
As a concession, an increase in the average training iterations when working with UIs can be seen in the last column. This is an expected effect of the MC sampling of the UIs and it amounts to rises of 6.26$\%$ and 3,72$\%$ respectively, which does not hinder the application of UI SVGPs when compared to DI SVGPs.
Finally, an expected increase in the SVGP variance (monitored through the kernel trace) is reported between SVGPs with DI and UIs. Nonetheless, when analysing the distribution of the sampled posterior variance over the mapped area, it can be seen that the variance values in the predicted area are in average lower for UIs. See Fig. \ref{fig:vars_gps} for an example.

\subsection{SVGP-PF on Predicted Terrain}
\begin{figure}[htbp]
    \centering
    \includegraphics[scale=0.5]{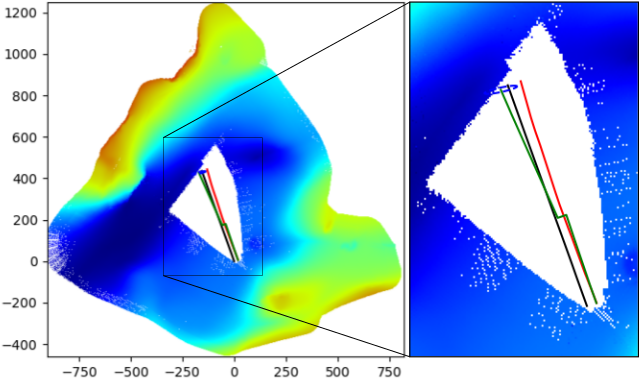}
    \caption{GT (black), DR (red) and PF (green) trajectories for one round of AUV localisation over unseen terrain}
    \label{fig:pf_result}
\end{figure}

An AUV localisation mission has been carried out over the second section of the survey, corresponding to unmmaped terrain.
The PF has been initialized approximately 20 m away from the real AUV pose in order to test if it could correct a DR drift equivalent to that accumulated through the first section of the survey.
Fig \ref{fig:pf_result} shows an instance of the experiment.
The Hugin has followed a straight trajectory of approximately 500 m (in black) over the unseen area. The DR trajectory is depicted in red and the PF result in green, with the final PF variance in blue.

For the experiments, a PF has been fine-tuned to achieve the best performance across the 5 UI maps generated on the previous section with yaw noise $1e^{-3} rad/s$. 
The mission setup and filter parameters have been kept the same and the measurement noise model has been again fine-tuned for best performance on DI maps.
The average results can be seen in Table \ref{tab:pf_results}.
The average RMS errors in position and heading across the full trajectories of the DR and the PFs are compared separately, as advised in \cite{olson2009evaluating}.
It can be seen that indeed the PF localisation in the UI maps yields a smaller error in the pose estimate with no significant difference in the heading.

The PF experiments were run with 60 particles at an average AUV speed of 2 knots per second (faster than the real AUV velocity in the mission) with additive Gaussian noise in the yaw with covariance $1e-^{4}$, on a laptop with an Intel processor i7-7700HQ CPU @ 2.80GHz and an NVIDIA GeForce GTX 1060 GPU.

\begin{table}[htbp]
\vspace*{0.1in}
\centering
\caption{PF localisation and DR errors against GT}
\begin{tabular}{|c|cc|}
\hline
 AUV localization & $RMSE_{xyz}$ & $RMSE_{\theta}$ \\ \hline
 DR         &  12.323 & 0.070 \\
 DI SVGP-PF & 11.882 &  0.064  \\
 UI SVGP-PF & 10.711 &  0.064  \\ \hline
\end{tabular}
\label{tab:pf_results}
\end{table}

\section{Conclusions}
We have presented a framework to learn SVGP maps with UIs, which harnesses the flexibility of Monte Carlo integration together with the capability of SVGP to handle very large datasets. 
When compared to the current state of the art method being used in autonomous mapping under localization uncertainty \cite{o2012gaussian, jadidi2015mutual, popovic2020informative}, this approach is not constrained to Gaussian UIs and it does not require to manually set any parameter for the UIs uncertainty approximation. Furthermore it is more straight-forward to implement, as it does not require modifying the GP formulation, and it scales better with the dimensionality of the input data.

In order to test this method in a real bathymetric surveying scenario with an AUV and a MBES sensor, we have implemented tools to propagate the AUV DR and sensor uncertainties to the UIs in an efficient manner based on \cite{barfoot2014associating}.
We have analyzed the gains and disadvantages of learning bathymetric maps with SVGP with DIs and UIs in both the tasks of terrain reconstruction and prediction. For different levels of noise in the dataset, the resulting UI SVGP maps consistently outperform the DI SVGP ones in the task of terrain prediction while showing similar results in terrain reconstruction. The concession is with regards to the training time, with a small increase in the number of required learning steps with UI SVGPs, as expected. Furthermore, although the resulting UI SVGP posterior variances are higher, an empirical analysis has shown that their values on the predicted area are on average lower, which correlates with the smaller prediction error.
Finally we present favourable results of the SVGP mapping technique with UIs against DIs in a real AUV localisation scenario with a PF in an entirely predicted area. These reinforce the conclusion that the UI SVGP has produced not only a more accurate prediction of the terrain but also of the variance in the area, which the PF has utilized to provide a more accurate pose estimate. 

Overall, the results of the work presented, although limited to one dataset, show the promising capabilities of the proposed method to apply UI SVGPs in the tasks of mapping, terrain prediction and vehicle localisation over DI SVGPs. 
Further research shall examine how this approach generalises to datasets containing different variations of the seabed topography and their full potential for PF localisation.
To facilitate this, we release our development setup in ROS for uncertainty propagation, SVGP mapping and AUV PF localisation.

\section*{Acknowledgement}
The authors thank the Alice Wallenberg foundation for funding MUST, Mobile Underwater System Tools, project that provided the Hugin AUV.
This work was supported by Stiftelsen fr Strategisk Forskning (SSF) through the Swedish Maritime Robotics Centre (SMaRC) (IRC15-0046).

\balance
\bibliography{main}
\bibliographystyle{unsrt}

\end{document}